\documentclass[review]{elsarticle}% 学术文章
\usepackage{amsmath}   %引入公式包
\usepackage{lineno,hyperref} % 引入宏包
\usepackage{algorithmic}
\usepackage{algorithm}
\usepackage{bbding}
\usepackage{pifont}
\usepackage{wasysym}
\usepackage{amssymb}
\usepackage{color}
\usepackage{amssymb}   
\usepackage{booktabs}   %三线表的宏包
\usepackage{multicol}  
\usepackage{multirow}  
\usepackage{caption}
\usepackage{graphicx}
\usepackage{diagbox}    %插表格用的宏包
\usepackage{multirow}   %插多行表格用的宏包
\usepackage{bm}  %用于公式加粗  %%\textbf{}用于加粗文本
\usepackage{subfigure} 

\modulolinenumbers[5]
\journal{Journal of \LaTeX\ Templates}   

\usepackage{graphicx}
\usepackage{caption}
\usepackage{subfigure} 
\usepackage{float}  
\usepackage[subfigure]{graphfig}
\usepackage{hyperref}
\hypersetup{hidelinks,
	colorlinks=true,
	allcolors=black,
	pdfstartview=Fit,
	breaklinks=true} 

\date{}
\begin{document}

\begin{frontmatter}

%\title{Triplet Center Loss based Part-aware Model for Vehicle Re-identification}  
 
\title{ Domain Adaptive Person Search via GAN-based Scene Synthesis for Cross-scene Videos 	}  

%Domain Adaptive Person Search via GAN-based Scene Synthesis for Cross-scene Videos 
%\author{Huibing Wang} %\corref{cor1} 
%\ead{huibing.wang@dlmu.edu.cn}address
%S
\author[first]{Huibing Wang}
\author[first]{Tianxiang Cui\corref{cor1}}
\author[first]{Mingze Yao\corref{cor1} }
\author[first]{Huijuan Pang }
\author[first]{Yushan Du }
\address[first]{School of Information Science and Technology, Dalian Maritime University, Dalian 116026, China}
\cortext[cor1]{Corresponding author.}

\begin{abstract}    %摘要

Person search has recently been a challenging task in the computer vision domain, which aims to search specific pedestrians from real cameras.Nevertheless, most surveillance videos comprise only a handful of images of each pedestrian, which often feature identical backgrounds and clothing. Hence, it is difficult to learn more discriminative features for person search in real scenes. To tackle this challenge, we draw on Generative Adversarial Networks (GAN) to synthesize data from surveillance videos. GAN has thrived in computer vision problems because it produces high-quality images efficiently. We merely alter the popular Fast R-CNN model, which is capable of processing videos and yielding accurate detection outcomes. In order to appropriately relieve the pressure brought by the two-stage model, we design an Assisted-Identity Query Module (AIDQ) to provide positive images for the behind part. Besides, the proposed novel GAN-based Scene Synthesis model that can synthesize high-quality cross-id person images for person search tasks. In order to facilitate the feature learning of the GAN-based Scene Synthesis model, we adopt an online learning strategy that collaboratively learns the synthesized images and original images. Extensive experiments on two widely used person search benchmarks, CUHK-SYSU and PRW, have shown that our method has achieved great performance, and the extensive ablation study further justifies our GAN-synthetic data can effectively increase the variability of the datasets and be more realistic. The code is available at \href{https://github.com/crsm424/DA-GSS}{https://github.com/crsm424/DA-GSS}
	
\end{abstract}

\end{frontmatter}

     %每五行一个标记   

%正文开始

\section{Introduction}   %一级标题

Person search aims to find a specific pedestrian from the given images or videos taken in real-world scenes, which is a challenging task in the recent computer vision domain. In recent years, person search has attracted increasing attention due to its practical application, such as smart surveillance systems \cite{8953988}, activity analysis \cite{6619065,wu2017robust}, people tracking in criminal investigations\cite{8555868,9229518}, and other fields. In general, existing person search methods adopt hand-cropped videos, which make the pedestrian bounding boxes clean and less noisy. However, hand-cropped processes require a lot of time and manpower, making them unsuitable for real-world scenarios. Therefore, person search needs to process the whole image which has a large number of pedestrians from the actual surveillance videos, rather than the pre-processed images. Besides, sharing features between detection and re-identification may also cause errors to accrue from each process, which will negatively impact the effectiveness of person search. The above two issues still cause challenges for existing person search methods to complete the task of real-time target searching in large-scale smart surveillance systems.

Deep learning-based methods for person search have proposed two different strategies to solve the above issues, which are named two-stage and one-stage respectively due to the framework differences. One-stage methods utilize the unified framework which combines person detection and person re-ID into an end-to-end model \cite{yan2021anchor,9911691,munjal2019query,chen2020norm}. These unified frameworks specifically advocate for an additional layer to be added behind the detection network in order to modify the person-bounding boxes for the re-identification network. They use a combined loss during training that consists of a person detection loss and a person categorization loss. The goals of the searching task, however, conflict with those of detection and re-ID, so the shared features between the two tasks are inappropriate. In other words, the detection and re-ID tasks aim to find the common features of pedestrians and the unique features of a specific person, respectively. Therefore, jointly learning the two tasks may influence the optimization of the model, and some researchers utilize a two-stage framework to separate them as two independent networks. Two-stage frameworks for person search \cite{9265450,han2019re,wang2020tcts,9003518} attempt to locate multiple pedestrians in the whole image with detection networks and then extract the above pedestrians which are fed to re-ID networks to complete re-identification task. Assisted in the great results of the detection model, two-stage frameworks mainly focus on how to effectively extract robust and discriminative representations.

Existing two-stage person search methods have achieved great performance, but they still fail to notice the contradictory requirements between the sub-tasks in person search. In detection networks, gallery images are auto-detected from the general detector, which produces a large number of bounding boxes for each pedestrian. As a result, the re-ID task's objectness information is ignored by the identified gallery images, which makes the problem of missed detection on query targets worse. Additionally, the re-ID framework does not agree with all of the detection results. Compared with existing re-ID datasets, the detected bounding boxes are more likely to have problems of misalignment, occlusions, and missing part of the person, even though the detected results do not contain person. Due to the aforementioned issues, the re-ID framework is unable to produce correct recognition results. The detection stage and re-ID stage consistency issue reduces search performance and restricts practicability. In order to resolve the contradiction between these two steps, we must therefore optimize the detection findings.    %%图中可能不完整的人 对re-id有影响

No matter whether researchers adopt a one-stage framework to jointly complete the two sub-tasks, or a two-stage framework to separately solve the two sub-tasks. It is believed that the accuracy of pedestrian detection and the retrieval performance of re-ID have a mutual influence. Note that while some of the aforementioned approaches have achieved great performance, we find that if the accuracy of the first part of pedestrian detection has been improved, the re-ID framework utilizes higher quality candidate samples to compare against the query, which can improve the performance of person search. Therefore, we consider it more effective to enhance the search performance by obtaining more pedestrian images. In real scenarios, including the wildly used two datasets for person search, the monitoring device is situated in diverse places, and the uncertain number of the samples from cameras may cause the pedestrians appearing in each image to be random, sparse, and unbalanced. In view of the above problems, some researchers propose image-operated \cite{9684399,wang2019color,yu2021apparel} methods for person re-identification and person search tasks, which aims to improve performance by generating diverse images. And the current video retrieval technology \cite{yang2021deconfounded,yang2022video,dong2021dual,qian2023adaptive} has been continually maturing, which greatly assists in the advancement of our two-stage video processing.   %%引入gan  提高re-id性能

%%我们提出了基于gan的什么什么方法用于行人搜索 ，，，

Hence, inspired by the above discussions, we proposed a novel approach for person search with generative adversarial networks to eliminate the contradiction between the detection and re-ID stage, termed as Domain Adaptive Person Search via GAN-based Scene Synthesis (DA-GSS) for cross-scene videos. It combines two stages: pedestrian detection and person re-identification. During the detection stage, an auxiliary identity query module (AIDQ) is devised to manage the video's detection results, aiming to crop the image and retain positive samples for the Reid stage. Specifically, the positive samples are obtained by discarding the background and unmarked identity images, and only keeping the instances that are likely to play an active role in the reid task. Besides, in order to enforce our model to learn more discriminative features, we adopt a generative adversarial network through scene synthesis in our proposed model.  Our proposed GAN-based scene synthesis model adopts a generative adversarial network to synthesize data for cross-scene videos, which can effectively generate high-quality images and overcome the challenge of real-scene person search. Specifically, the proposed GAN-based scene synthesis model adopts encoders to separate appearance information and structure information from person images and utilizes decoders to synthesize person images with different appearance information.  Moreover, we also design a discriminative module in the above model, which aims to online learn discriminative features from the synthetic images and original images. In summary, the major contributions of the proposed method are the following:
 %% gan的作用 + 怎么用的，，， 

\begin{itemize}
	\item We propose a framework for domain adaptive person search that utilizes GAN-based scene synthesis for cross-scene videos, which can synthesize high-quality images across different videos, and learns discriminative features for person search.  
	\item In order to relieve the pressure bought by the two-stage model, we design an Assisted-Identity Query module for cropping the person image from the whole image and providing positive images for the behind part that can improve the overall performance of the person search model.
	\item To make the proposed model more discriminative and robust for person search, GAN is used to synthesize cross-scene person data, which enforces our network to learn more discriminatory and finer-grained features. Besides, we conduct experiments on the widely used CUHK-SYSU and PRW datasets and find that the newly synthesized data helps improve the performance of the model.
\end{itemize}

The remainder of the paper is outlined as follows. Section 2 introduces the related work. Section 3 presents the proposed methods about domain adaptive person search. Extensive experiments including complexity analysis and ablation study are conducted to verify our proposed model in Section 4. Finally, Section 5 concludes this paper.

\section{Related Work}
We first review the existing works on person search, which have drawn much interest recently. We also review some recent works about the two fields: generative adversarial network and person re-identification, which are the components of our proposed framework.

\subsection{Person Search}
Person search has increasingly become popular and studied since the publication of two large-scale datasets, CUHK-SYSU \cite{xiao2017joint} and PRW \cite{zheng2017person}. Recently, most of the research work takes end-to-end models into consideration. Xiao et al. \cite{xiao2017joint} propose the first one-stage person search model, which is trained with an online instance matching loss function. Xiao et al. \cite{xiao2019ian} design an Individual Aggregation Network (IAN) and introduce a center loss to increase the intra-class compactness of feature representations. Yan et al. \cite{yan2019learning} exploit contextual information to improve the discriminativeness of the learned features. Chen et al. \cite{chen2020norm} present an embedding decomposing method to deal with the contradictory objective problem of person search. Yan et al. \cite{yan2021anchor} propose a Feature-Aligned Person Search Network to tackle the problems of scale, region, and task misalignment. Li et al. \cite{li2021sequential} replace the bounding boxes with low-quality proposals.

Besides the end-to-end networks, other works solve person search in two stages i.e. training two parameter independent models for detection and re-ID. Chen et al. \cite{chen2018person} adopt a Mask-Guided Two-Stream (MGTS) method, which extracts more features by separating the detector and re-ID. Han et al. \cite{han2019re} introduce an ROI transform layer, which provides the refined detection boxes for person search. Wang et al. \cite{wang2020tcts} propose a Task-Consistent Two-Stage (TCTS) framework to deal with the questions of inconsistency. Especially, Yan et al. \cite{yan2022exploring} only employ the bounding box annotations, opening the door to a new world of weakly supervised and even unsupervised person search.

\subsection{Person Re-identification}
Due to the rapid development of intelligent monitoring systems, re-identification tasks have received increased attention in recent years. Early person re-ID method mainly focus on hand-crafted features \cite{7298832, WANG202155,5539926,liu2022delving,zeng2020energy,SHI2022104335} and learning distance metrics \cite{7410777,7780512,9140403,wang2021survey,yang2017person,LV2020103875}. For example, Liao et al. \cite{7298832} propose to extract features with HSV color histogram and SILTP descriptors, which can make a stable representation against viewpoint changes. Zhang et al. \cite{7780512} present an LSSCDL algorithm to learn a pair of dictionaries and a mapping function efficiently, which adopt a sample-specific SVM for each person. With the rise of deep learning, CNN-based models have attracted more attention \cite{6909421,9106791,xu2021rethinking,wang2022progressive}. Many approaches learn global features from the whole images or videos directly. However, these methods typically suffer from misalignment problems, occlusion problems, and background interference problems. Song et al. \cite{song2018mask} construct synthetic RGB-Mask by introducing the binary segmentation masks and designing a mask-guided attention model to learn features from each part of the whole image. Ye et al. \cite{ye2016person} propose a ranking aggregation algorithm to enhance the similarity information in detection, which can achieve great performance in video datasets. Moreover, some researchers have proposed person re-identification with synthetic data by computer. Sun et al. \cite{sun2019dissecting} introduce a large-scale synthetic dataset named PersonX which is composed of hand-crafted 3D person models. They synthesize pedestrians in a controllable manner. Chen et al. \cite{chen2019instance} propose a novel instance-guided context rendering scheme to transfer the source domain with context information. In summary, synthetic data for person re-identification tasks have captured the attention of many scientific researchers.

\subsection{Generative Adversarial Network}
The method of generating diverse realistic images has already been a hot-spots and cutting-edge direction because deep learning-based models need large-scale samples for training. Therefore, the generative adversarial network has received more interest than before. Wang et al. \cite{wang2019color} propose a novel color-sensitive network for person re-identification tasks, which utilizes a color translation method to generate fake images with different clothing colors. The generated images can effectively improve the complexity of training data, which can improve the performance of the deep learning-based model. Yu et al.\cite{yu2021apparel} propose a semi-supervised learning framework to generate person images with different clothes, which aims to learn apparel-invariant features from different pedestrian representations. Furthermore, Eom et al.\cite{eom2021gan} have introduced a new generative adversarial network that can factorize person image into identity-related and identity-unrelated features. The authors divide the whole person image into different parts(e.g., clothing, human pose, background), and utilize the proposed network to verify the identity-related features. Note that while the above researches have achieved great performance, they mainly focus on the appearance gap between the different persons and ignore structure information in person images. Chen et al.\cite{chen2021joint} have designed a 3D mesh rotation strategy to generate different view images for the same person. They also introduce a view-invariant loss to verify the influence of different positions and facilitate contrastive learning between original images and generated images. Yao et al.\cite{yao2020gan} have proposed a novel framework for person search tasks. They utilize a generative adversarial network to generate person images with different backgrounds, which solves the typical problem that low numbers of each person in person search datasets. Inspired by the above work, we generate different images with human appearance and structure information and effectively solve the problem of the lack of each person images in datasets.

\begin{figure}[tbp!]
	\centering
	\includegraphics[width=\textwidth]{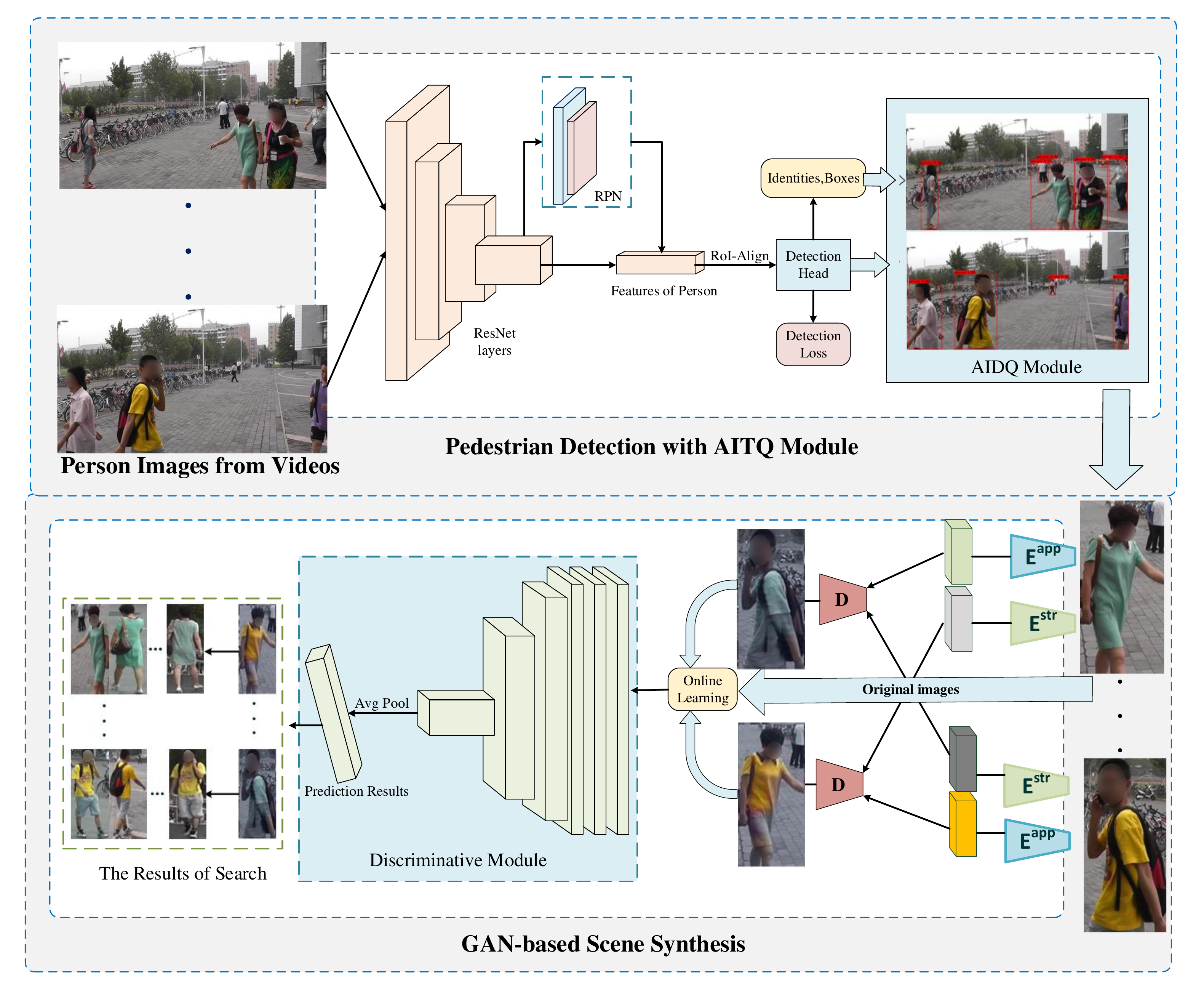} 
	\caption{The Whole Process of Our proposed framework}
	\label{whole}
\end{figure}

 \section{Domain Adaptive Person Search Method}
 As illustrated in Fig. \ref{whole}, we show the details of our proposed person search framework DA-GSS, which can be generally divided into two parts, the detection model with AIDQ and the GAN-based Scene Synthesis network. In this section, we first present an overview of our whole frame and then describe more details for the proposed DA-GSS.

 \subsection{overview}
 A panoramic query image is first input into the pedestrian detector, which outputs several proposals with their confidence scores. We suppress those instance bounding boxes whose confidence scores are lower than the given threshold and send the remaining ones through ROI-Align alignment to the detection head for final prediction results. An Assisted-Identity Query (AIDQ) module is used to separate labeled and unlabeled identities in the results and to crop the labeled identities into the GAN-based Scene Synthesis network.
 
 After that, pairs of cropped images of different identities are fed into the adversarial generative network. Our proposed DA-GSS model introduces a generative module that decomposes pedestrian images into two parts: appearance information, which mainly contains appearance semantics and other identity-related information, and structure information, which consists of geometry and position-related structural information, along with other variations. Therefore, a new image can be generated by combining the two parts of information from different people with each other. What's more, the synthesized image is given a soft label with a teacher model. Finally, the input images and the outputs image are jointly fed into the discriminant module to realize the person search task.
 
 The pedestrian detector and GAN-based Scene Synthesis model are trained independently. And we train the GAN-based Scene Synthesis model using ground truth annotations instead of detections to avoid errors caused by the detector.

 \subsection{Pedestrian Detection with AIDQ module}
 As shown in Fig. \ref{whole}, our detection model based on Faster R-CNN \cite{ren2015faster} is composed of a base network named ResNet50 \cite{he2016deep} for feature extraction, a region proposal network (RPN) for proposal generation, a classification network for final predictions and an Assisted-Identity Query (AIDQ) module for separating and cropping identities.
 
 In previous research work, both the two-stage and end-to-end models generally use the instance bounding boxes to perform the re-id task. Although the number of proposals is large enough to learn better features, most of the instance bounding boxes are of poor quality and can not learn fine-grained and discriminative features. Therefore, in order to solve the first problem, inspired with the idea of SeqNet \cite{li2021sequential} to use the final detection frame to perform the re-id task. The AIDQ module is proposed to produce high-quality positive samples for the re-id stage. The AIDQ module is trained by a classification loss called AIDQ loss in order to introduce the identity information into the detector. In this way, the AIDQ module can compute identity similarity scores between query target and detection results.

 Specifically, the DA-GSS model is proposed that reads each frame of the query video into the detector and then generates candidate frames after the backbone and RPN. In order to obtain accurate detection results from the input videos, our proposed model designs an NMS suppression with a threshold of 0.5 (CUHK)/0.6 (PRW) before the detection head, which aims to drop the negative results from the detection network. Finally, the proposed AIDQ crops out the detection images with labeled identity and fed them to the GAN-based Synthesis model, which can better alleviate the two tasks. %%没改
 
 It is worth discussing how to design a suitable loss function for it. The aim of AIDQ is to output a cropped image that is closer to the labeled identity ground truth than other bounding boxes. Therefore, it has some differences from the traditional classification or re-ID task. Next, we first introduce two common classification losses. Then, we derive our proposed AIDQ loss.
 
 One of the most well-known losses is the softmax loss, which is frequently employed in picture segmentation and classification tasks. It is composed of softmax and cross-entropy loss. For an example, $x_i$, the probability of $x_i$ being recognized as class $i$ is, Softmax loss uses a Softmax function to calculate the probability on each class:
 \begin{equation}
 	p_i = Softmax(z_i) = \frac{e^z_i}{\sum_{c=1}^{C}e^z_c},
 \end{equation}
 where $z_i$ is the output value of the $i$th node, and $C$ is the number of categories of classification. Then use a cross-entropy loss function to optimize the log-likelihood of each class in probability space:
 \begin{equation}
 loss_i = -log\left(p_i\right),
 \end{equation}
 
 OIM loss is first proposed for person search task in \cite{xiao2017joint}. Different from the Softmax loss, the OIM loss stores a feature center for each person. Specifically, the re-id features of all the training instances are stored in a memory. For each feature $x_i$:
 \begin{equation}
 L_i = -log\frac{exp(x_i\cdot m^+/\tau)}{\sum_{j=1}^{N_c}exp(x_i\cdot m_j/\tau)},
 \end{equation}
 where $m^+$ = $m_j$ if $x_i$ belongs to the $j$-th cluster, '$\cdot$' denotes the inner product, and $\tau > 0$ is a temperature hyper-parameter that controls the softness of the probability distribution.
 
 The proposed model have not directly adopted these classification losses for two reasons. For one thing, traditional classification loss may not be the best solution in pedestrian search tasks and the unlabeled examples are not fully exploited in these two losses. The circular queue length is a parameter that is artificial, even though OIM loss takes unlabeled identities into account. If the length is too large, the primitive features in the circular queue are outdated to represent unlabeled identities. If small, the optimizing direction and solution are changed significantly in different mini-batch. For another thing, positive and negative samples also contribute equally to these two losses. However, we are more inclined to believe that increasing the model's capacity for discrimination requires paying closer attention to hard samples.

As indicated by prior work TCTS \cite{wang2020tcts}, its proposed AIDQ loss learns a variable number of centers for unlabeled samples, which successfully solves the first problem. The proposed model introduce an AIDQ loss to improve the ability to select high-quality positive samples to better solve the consistency problem between the detection and re-ID tasks. For each labeled example, the AIDQ loss pulls the positive examples from the different images closer, so that the images from the same people can receive a high similarity score. Additionally, the AIDQ loss distances the example from the negative instances (including unlabeled examples) in the same images, reducing the similarity between various individuals. We also fully consider the positive role of hard negative samples in improving the discriminative ability of the network. Without loss of generality, all negative samples are sorted in descending order of score$\left\{c_1^-,...,c_N^-\right\}$, where the number of negative samples is $N^- = N_c-1$. We set the number of hard negative samples with:
 \begin{equation}
\label{4}
K = \mathop {\arg \min }\limits_k \left|\frac{\sum_{m=1}^{k}x_i\cdot c_m^-}{\sum_{n=1}^{N^-}x_i\cdot c_n^-}-\lambda \right|,
 \end{equation} 
 where $\lambda$ is a threshold that controls the ratio of hard negative samples. Therefore, the AIDQ loss assuming a total of $N$ samples is defined as:
 \begin{equation}
L_{AIDQ} = -\frac{1}{N}\sum\limits_N log\frac{exp\left(x_i\cdot c^+/\tau\right)}{exp\left(x_i\cdot c^+/\tau\right)+\sum_{k=1}^{K}exp\left(x_i\cdot c_k^-/\tau\right)},
 \end{equation}

\subsection{GAN-based Scene Synthesis}
At this stage, the proposed DA-GSS model design a generative adversarial network (GAN) to synthesize data in videos. Since the datasets we utilize are extracted from video frames, which make samples similar and have a low number of samples in big data problem. DA-GSS integrated GAN and re-identification into a unified model which can dynamically generate images and complete person re-ID tasks with new images. Inspired by DG-Net \cite{zheng2019joint}, we design a generative module to generate high-quality samples from videos and a discriminative module to complete the re-identification task. Specifically, the proposed generative module utilizes an encoder-decoder paradigm, which can effectively extract features from input data. There are two types of input data in the GSS module: appearance information, which has mostly apparel and identity-related semantics, and structure information, which has geometry and positional information. We refer to the extracted features from the encoder as "code". Besides, our proposed discriminative module is collaborated learning with the generative module, which utilizes an "online learning" manner. In a word, the generated images are fed into the discriminative module directly, and the two modules are sharing the encode layer structure with each other.  

In our re-ID network, we denote the detection result images and identity labels as $X=\{x_i\}_{i=1}^N$ and $Y=\{y_i\}_{i=1}^N$, respectively. $N$ is the number of the input images, $y \in [1,M]$ and $M$ indicates the number of person identities in the dataset. As shown in Fig.\ref{whole}, the result images from the AIDQ module are fed to our designed unified model in pairs. In our proposed model, given two person images $x_i$ and $x_j$, they are divided into appearance information with encoding layer as $E_{app}:x_i \to c_i^{app} $, and structure information is similarly extracted as $E_{str}:x_j \to c_j^{str} $. Meanwhile, the decoding layer aims to generate new images with the above information from different person, which can be shown as $D: (c_i^{app},c_j^{str}) \to x_{ji}$. Overall, our generative model utilizes an encoder-decoder paradigm to synthesize a new person image by exchanging the appearance codes and structure codes from the input two pictures. Note: for the generated image, we defined it as $x_{ji}$, in which $j$ indicates the one offering structure code and $i$ denotes the person image providing appearance code. As illustrated in Fig.\ref{whole}, we utilize the encoder-decoder paradigm to synthesize new person images with different appearance codes, which enforce the network to mine more fine-gained features rather than explicit appearance features. Given two images $x_i$ and $x_j$ of different identities $y_i \ne y_j$, the generative module learns how to synthesize the same person with a different cloth. The synthetic image $x_{ji}=D(c_i^{app},c_j^{str})$ is required to contain the information of appearance code $c_i^{app}$ from $x_i$ and structure code $c_j^{str}$ from $x_j$, respectively. In order to prove the two latent codes to be reconstructed after synthesizing images, we using the pixel-wise $l_1$ loss:
\begin{align}
L_{recon}^{app} =& \mathbb{E}[||c_i^{app} - {E_{app}}(D(c_i^{app},c_j^{str}))|{|_1}],\\
L_{recon}^{str} =& \mathbb{E}[||c_i^{str} \ - {E_{str}}(D(c_i^{app},c_j^{str}))|{|_1}].
\end{align}
where $\mathbb{E}[||\cdot|{|_1}]]$ means $l_1$ loss function, and we also propose a identification loss to enforce the synthetic image to keep the identity consistency:
\begin{equation}
\label{id}
L_{id}^{s} = \mathbb{E}[ - \log (p({y_i}|{x_{ji}}))|{|_1}]
\end{equation}
where $p({y_i}|{x_{ji}})$ is the predicted probability of $x_{ji}$ belonging to the ground-truth class $y_i$ of $x_i$, which provides appearance code in synthesizing $x_{ji}$. Moreover, we adopt adversarial loss like normal GAN networks to match the distribution of synthetic images to the real data distribution:
\begin{equation}
{L_{adv}} = \mathbb{E}[\log F({x_i}) + \log (1 - F(D(c_i^{app},c_j^{str})))]
\end{equation}
where $F(\cdot)$ presents the function of distribution.

In our proposed discriminative module, we first utilize a teacher-guided model to generate soft labels for synthetic images from the generative module \cite{qian2022switchable}. For normalization, we adopt PCB \cite{sun2018beyond} as the teacher network to assign a soft label for the synthetic image $x_{ji}$, and we just simply train the teacher model on the original training set. In order to effectively learn the synthetic person images, we utilize KL divergence to minimize the probability distribution $p(x_{ji})$ predicted by the discriminative module and the probability distribution $q(x_{ji})$ predicted by the teacher:
\begin{equation}
{L_{KL}} = \mathbb{E}[ - \sum\limits_{m = 1}^M {q(m|{x_{ji}})\log (\frac{{p(m|{x_{ji}})}}{{q(m|{x_{ji}})}})} ]
\end{equation}
where $M$ is the number of identities.

Besides, our discriminative module utilizes discriminative feature extracted layers to complete the person re-identification task. Specifically, in accordance with the synthetic person images by switching appearance and structure codes, we propose discriminative and fine-gained features mining to better take advantage of the collaborated online learning. 
Our discriminative module is compelled to learn the fine-gained identity-related qualities (such as hair, face, bag, and other traits) that are independent of clothes during training on the synthetic images. We synthesize the person images by one structure code combined with other appearance codes, which are regarded as the same class as the real image based on the structure code. In order to achieve the above function, we enforce identification loss on this specific categorizing:
\begin{equation}
L_{loc} = \mathbb{E}[ - \log (p({y_i}|{x_{ji}}))|{|_1}]
\end{equation}

This loss is similar to Eq.\ref{id} but has a different meaning. We adopt this loss function to impose identity supervision on the discriminative module in a multi-tasking way. Moreover, unlike the previous approaches using manually partitioned person images to learn discriminative features \cite{sun2018beyond,zhao2021learning,shen2019part}, our approach performs automatic discriminative features learning by leveraging on the synthetic images. As a result, our proposed discriminative module learns to attention to the identity properties through discriminative feature learning.
%%鉴别模块

In the widely used datasets, person images from videos are mostly appearance-invariant. But in practical application, pedestrians may change their clothes, which makes the network not suitable for the re-id task. Therefore, our synthetic images recompose the visual contents from real data. We regard the high-quality synthetic person image as "inliers", which can provide a positive effect on network learning and complete person re-identification task effectively. Besides synthesizing person images, our proposed GAN-based person re-identification network can also learn discriminative features for person re-identification using the two modules described above. Furthermore, our generative module synthesizes data using appearance code and structural code. In theory, we can synthesize $N \times N$ different images in an online generated training sample pool rather than ones with $2 \times N$ images offline generated in \cite{9157071,chen2019instance,tang2020cgan,wang2016iterative}

\begin{figure*}[tbp!]
	\centering
	\setlength{\belowcaptionskip}{-1mm}
	\vspace{-0.35cm} %设置与上面正文的距离
	\subfigtopskip=-1pt %设置子图与上面正文或别的内容的距离
	\subfigbottomskip=-1pt %设置第二行子图与第一行子图的距离，即下面的头与上面的脚的距离
	\subfigcapskip=-5pt %设置子图与子标题之间的距离
	\subfigure[mAP performance]{
		\includegraphics[width=5cm]{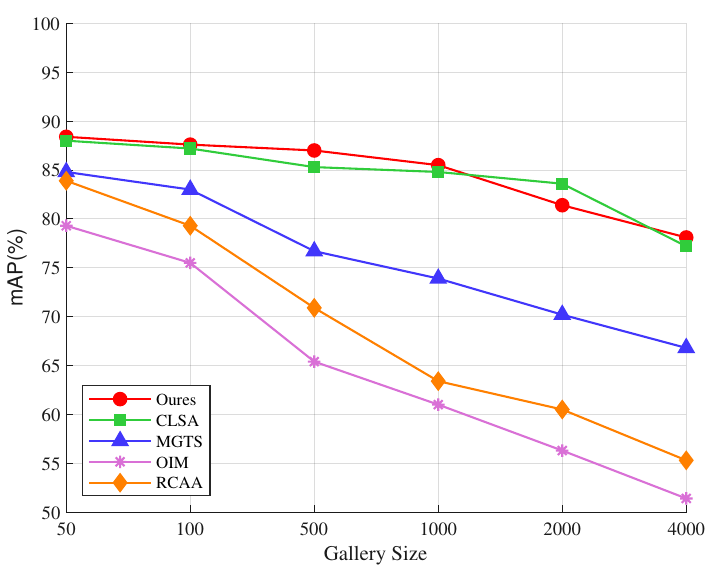}}
	\subfigure[CMC TOP-1 performance]{
		\includegraphics[width=5cm]{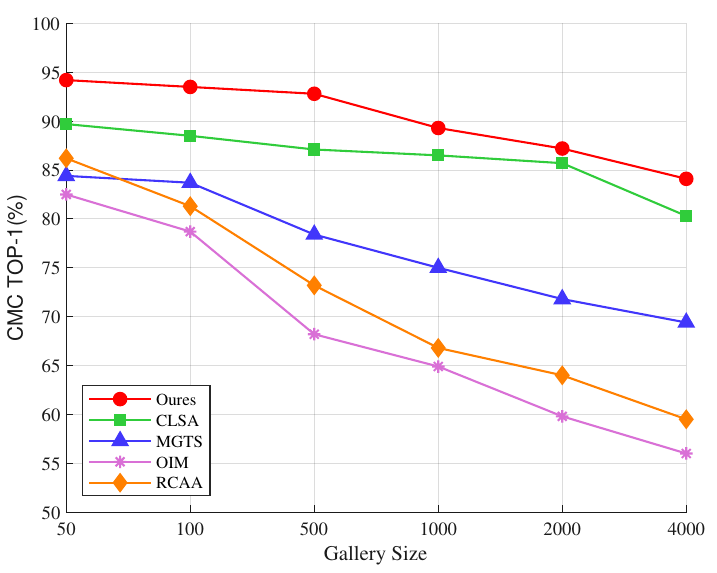}}			
	\caption{Comparison of person search performance of various gallery sizes on CUHK-SYSU dataset. }
	\label{GS}
\end{figure*}

\section{Experiments}

In this part, we conduct experiments on the two benchmark datasets, CUHK and PRW. Besides, we also adopt ablation study on different part of our proposed model.

\subsection{Datasets}
CUHK-SYSU \cite{xiao2017joint} is a large-scale dataset designed for person search, which contains 18,184 scene images captured from street nap and movie screenshot. Besides, 96,143 pedestrian bounding box annotations and 8,432 identities are marked in total. All people are divided into 8,432 labeled identities and other unknown ones. The train set contains 11,206 images with 5,532 different identities.
The test set contains 6,978 images with 2,900 query people. For each query, different gallery sizes from 50 to 4000 are pre-defined to evaluate the search performance. If not specify, gallery size of 100 is used by default.

PRW \cite{zheng2017person}, containing 11,816 video frames with 34,304 pedestrian BBoxes and 932 identity labels, is another person search dataset. All images are captured 6 static cameras in Tsinghua university and annotated manually. The training set includes 5,704 pictures and 482 labeled identities, while the test set has 6,112 images with 2,057 query people and 19,124 bounding boxes.
% Please add the following required packages to your document preamble:
% \usepackage{multirow}
\begin{figure*}[tbp!]
	\centering
	\setlength{\belowcaptionskip}{-1mm}
	\vspace{-0.35cm} %设置与上面正文的距离
	\subfigtopskip=-1pt %设置子图与上面正文或别的内容的距离
	\subfigbottomskip=-1pt %设置第二行子图与第一行子图的距离，即下面的头与上面的脚的距离
	\subfigcapskip=-5pt %设置子图与子标题之间的距离
	\subfigure[mAP performance]{
		\includegraphics[width=5cm]{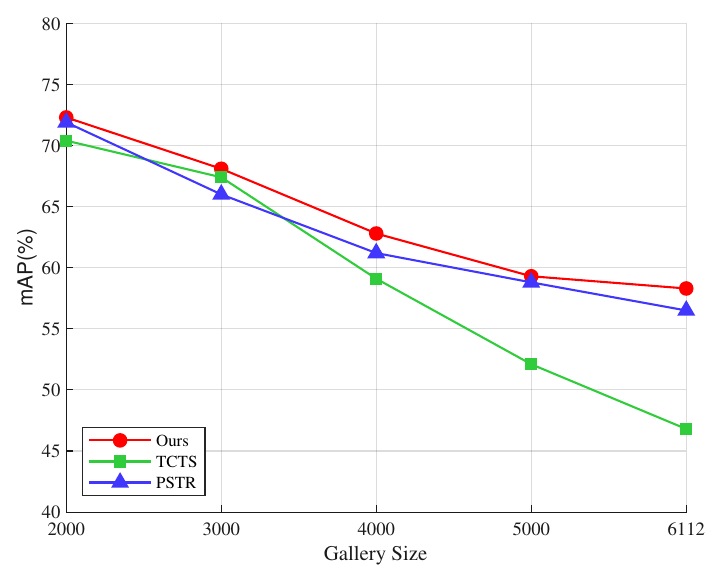}}
	\subfigure[CMC TOP-1 performance]{
		\includegraphics[width=5cm]{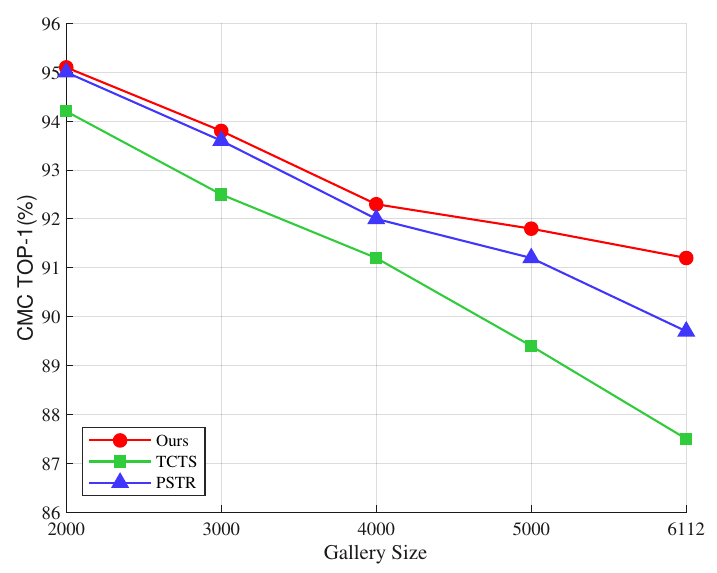}}			
	\caption{Comparison of person search performance of various gallery sizes on PRW dataset. }
	\label{PR}
\end{figure*}

\subsection{Evaluation Protocols}
With the presentation of \cite{munjal2019query,chen2020norm}, the Cumulative Matching Characteristic (CMC) and the mean Averaged Precision (mAP) are employed as the performance metrics. The average precision for each query $q$ is calculated by:
\begin{equation}
AP(q) = \frac{{\sum\limits_{k = 1}^n {{P_{(k)}}}  \times rel(k)}}{{{N_{gt}}}},
\end{equation}
where $P_{(k)}$ represents the precision of the $k_{th}$ position of the result. If the $k_{th}$ result matches correctly the $rel(k)$ is an indicator function equal to 1 or zero otherwise. $n$ is the number of tests and $N_{gt}$ is the ground truth. After experimenting with each query image, mAP is calculated as follows:
\begin{equation}
mAP = \frac{{\sum\limits_{q = 1}^Q AP(q) }}{Q},
\end{equation}
where $Q$ is the number of all queries. Therefore, we also adopt above standard metrics for person search performance evaluation.
\begin{table}[tbp]
	\centering
	\caption{Comparison of mAP($\%$) and rank-1 accuracy($\%$) with the state-of-the-art on CUHK-SYSU and PRW} 
	\label{Table1}
	\resizebox{0.7\textwidth}{6cm}{
		\begin{tabular}{cl|cc|cc}
			\hline
			\multicolumn{2}{c}{\multirow{2}{*}{Method}}                    & \multicolumn{2}{c}{CUHK-SYSU} & \multicolumn{2}{c}{PRW} \\ \cline{3-6} 
			\multicolumn{2}{c}{}                                           & mAP           & top-1          & mAP        & top-1       \\ \hline
			\multicolumn{1}{c}{\multirow{6}{*}{\rotatebox{90}{two-stage}}}   & DPM \cite{girshick2015deformable}         & -             & -              & 20.5       & 48.3        \\
			\multicolumn{1}{c}{}                             & MGTS \cite{chen2018person}        & 83.0          & 83.7           & 32.6       & 72.1        \\
			\multicolumn{1}{c}{}                             & CLSA \cite{lan2018person}        & 87.2          & 88.5           & 38.7       & 65.0        \\
			\multicolumn{1}{c}{}                             & IGPN \cite{dong2020instance}       & 90.3          & 91.4           & 47.2       & 87.0        \\
			\multicolumn{1}{c}{}                             & RDLR \cite{han2019re}       & 93.0          & 94.2           & 42.9       & 70.2        \\
			\multicolumn{1}{c}{}                             & TCTS \cite{wang2020tcts}       & 93.9          & 95.1           & 46.8       & 87.5        \\ \hline
			\multicolumn{1}{c}{\multirow{16}{*}{\rotatebox{90}{end-to-end}}} & OIM \cite{xiao2017joint}        & 75.5          & 78.7           & 21.3       & 49.4        \\
			\multicolumn{1}{c}{}                             & IAN \cite{xiao2019ian}         & 76.3          & 80.1           & 23.0       & 61.9        \\
			\multicolumn{1}{c}{}                             & NPSM \cite{liu2017neural}        & 77.9          & 81.2           & 24.2       & 53.1        \\
			\multicolumn{1}{c}{}                             & RCAA \cite{chang2018rcaa}       & 79.3          & 81.3           & -          & -           \\
			\multicolumn{1}{c}{}                             & CTXG \cite{yan2019learning}       & 84.1          & 86.5           & 33.4       & 73.6        \\
			\multicolumn{1}{c}{}                             & QEEPS \cite{munjal2019query}      & 88.9          & 89.1           & 37.1       & 76.7        \\
			\multicolumn{1}{c}{}                             & APNet \cite{zhong2020robust}      & 88.9          & 89.3           & 41.9       & 81.4        \\
			\multicolumn{1}{c}{}                             & HOIM \cite{chen2020hierarchical}       & 89.7          & 90.8           & 39.8       & 80.4        \\
			\multicolumn{1}{c}{}                             & NAE \cite{chen2020norm}        & 91.5          & 92.4           & 43.3       & 80.9        \\
			\multicolumn{1}{c}{}                             & NAE+ \cite{chen2020norm}       & 92.1          & 92.9           & 44.0       & 81.1        \\
			\multicolumn{1}{c}{}                             & AlignPS \cite{yan2021anchor}    & 94.0          & 94.5           & 46.1       & 82.1        \\
			\multicolumn{1}{c}{}                             & SeqNet \cite{li2021sequential}     & 93.8          & 94.6           & 46.7       & 83.4        \\
			\multicolumn{1}{c}{}                             & SeqNet+CBGM \cite{li2021sequential} & 94.8          & 95.7           & 47.6       & 87.6        \\
			\multicolumn{1}{c}{}                             & COAT \cite{yu2022cascade}       & 94.2          & 94.7           & 53.3       & 87.4        \\
			\multicolumn{1}{c}{}                             & COAT+CBGM \cite{yu2022cascade}  & 94.8          & 95.2           & 54.0       & 89.1        \\
			\multicolumn{1}{c}{}                             & PSTR \cite{cao2022pstr}       & 95.2          & 96.2           & 56.5       & 89.7        \\ \hline
			\multicolumn{1}{c}{Ours}                         & DA-GSS        & 87.6          & 93.5           & 58.3       & 91.2        \\ \hline
		\end{tabular}
	}
\end{table}

\subsection{Implementation Details}
We run all experiments on one NVIDIA GeForce 3090 GPU and achieve our model with PyTorch \cite{paszke2017automatic}. On the detection training model, the proposed model adopt Faster-RCNN \cite{ren2015faster} base on ResNet-50 \cite{he2016deep} pre-trained
on ImageNet \cite{deng2009imagenet}. During the training of the detection model, the batch size is set to 8 and each image is resized to 600$\times$800 pixels. The proposed DA-GSS model uses SGD optimizer with momentum for ResNet models, in which starting learning rate is set to $e^{-3}$ and decays to $e^{-5}$. Our proposed AIDQ module is built upon conv4 and adopts conv5 in ResNet-50. We pass the ground-truth box into the detector to assist the AIDQ module to learn and crop the positive samples. The proposed model trains the detection model for 40 epochs, reducing the learning rate by a factor of 10 at epochs 15 and 30. For the GAN-based Scene Synthesis model, (1) $E_{app}$ is also based on ResNet-50 pre-trained on ImageNet, and we remove its global average pooling layer and fully-connected layer. In order to obtain the appearance code $c^{app}$  in 2048$\times$4$\times$1, we append an adaptive max pooling layer behind. (2) $E_{str}$ is a shallow network that outputs the structure code $c^{str}$ in 128$\times$64$\times$32. It consists of four convolutional layers followed by four residual blocks \cite{he2016deep}. (3) Decoder $D$ processes $c^{str}$ by four residual blocks and four convolutional layers. As in \cite{huang2018multimodal} every residual block contains two adaptive instance normalization layers, which integrate into $c^{app}$ as scale and bias parameters. (4) SGD is used to train $E_{app}$ with learning rate 0.003 and momentum 0.9. We apply Adam to optimize $E_{str}$, $D$ and online learning module, and set the learning rate to 0.0001, and ($\beta_1$, $\beta_2$) = (0, 0.999). We train for a total of 100,000 iterations.

\begin{figure*}[tbp!]
	\centering
	\setlength{\belowcaptionskip}{-1mm}
	\vspace{-0.35cm} %设置与上面正文的距离
	\subfigtopskip=-1pt %设置子图与上面正文或别的内容的距离
	\subfigbottomskip=-1pt %设置第二行子图与第一行子图的距离，即下面的头与上面的脚的距离
	\subfigcapskip=-5pt %设置子图与子标题之间的距离
	\subfigure[CUHK-SYSU dataset]{
		\includegraphics[width=5cm]{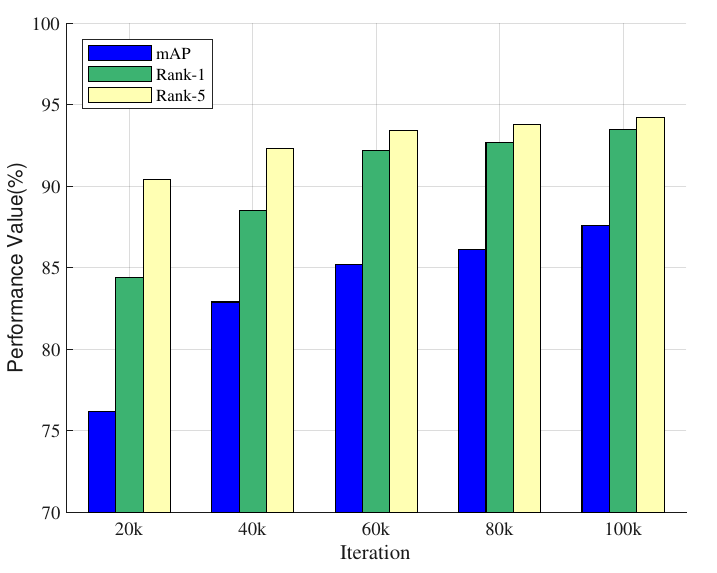}}
	\subfigure[PRW dataset]{
		\includegraphics[width=5cm]{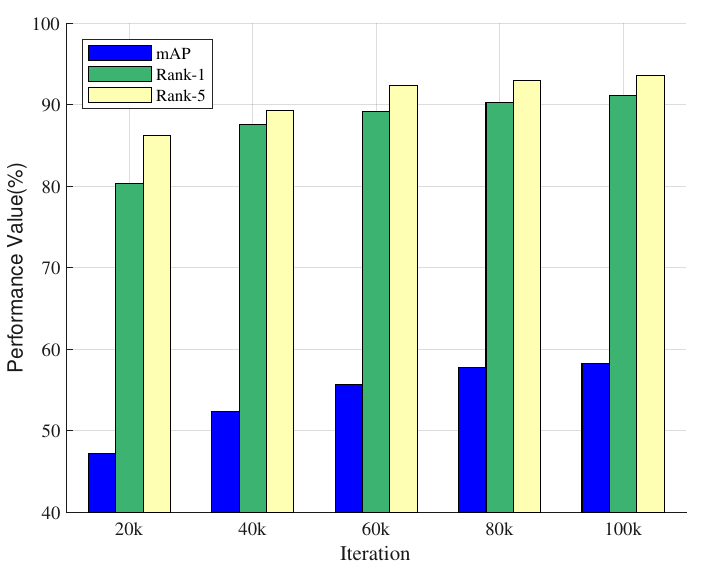}}			
	\caption{Comparison of person search performance of different iterations on CUHK-SYSU and PRW datasets. }
	\label{it}
\end{figure*}

\subsection{Comparison with State-of-the-art Methods}
We display a comparison of our proposed DA-GSS framework with several state-of-the-art methods on the standard benchmarks in Table \ref{Table1}. Although the performance of our DA-GSS on the CUHK-SYSU dataset is not eye-catching, which is still a long way from the current optimal results, it is beneficial to all indicators on the PRW dataset. Most importantly, we are the first framework to attempt to implement Generative Adversarial Networks inside the whole framework and achieve very good results. Compared to the previous best PSTR model, we achieve better performance on the PRW dataset.

\begin{figure*}[tbp!]
	\centering
	\setlength{\belowcaptionskip}{-1mm}
	\vspace{-0.35cm} %设置与上面正文的距离
	\subfigtopskip=-1pt %设置子图与上面正文或别的内容的距离
	\subfigbottomskip=-1pt %设置第二行子图与第一行子图的距离，即下面的头与上面的脚的距离
	\subfigcapskip=-5pt %设置子图与子标题之间的距离
	\subfigure[CUHK-SYSU dataset]{
		\includegraphics[width=5cm]{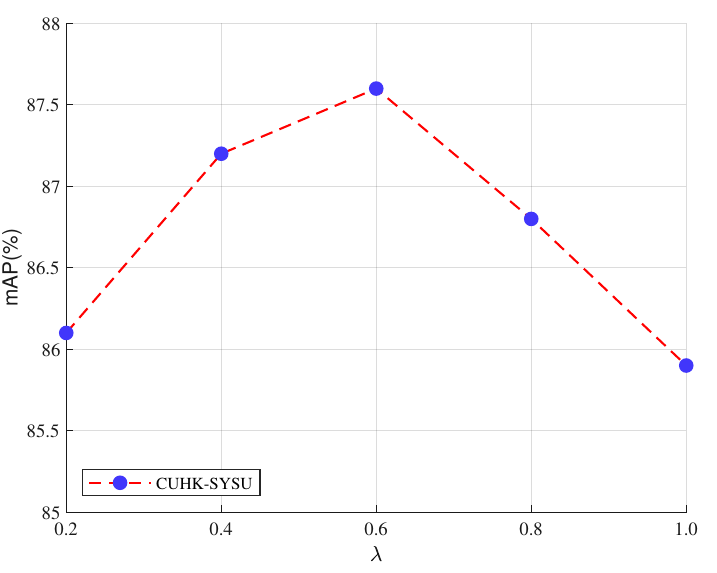}}
	\subfigure[PRW dataset]{
		\includegraphics[width=5cm]{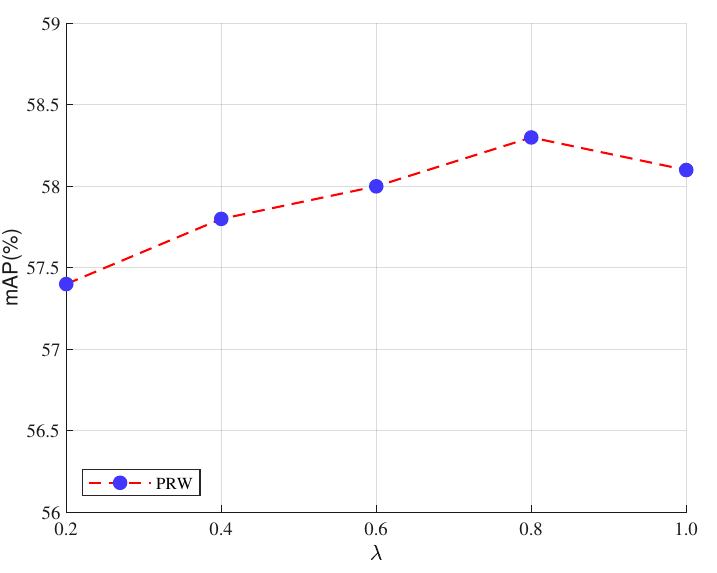}}			
	\caption{The mAP accuracy of the person search for the CUHK-SYSU dataset and PRW dataset  verification set using different weights $\lambda$. }
	\label{lambda}
\end{figure*}

This is a very surprising finding that generative adversarial network performs well on the real scene. As we all know, CUHK includes real scenes and movie footage, while PRW only includes pictures from real cameras. Based on this, we analyze that our proposed DA-GSS may be more suitable for real scenarios. In addition, the task of person search becomes more challenging when the gallery size increases violently. We vary the gallery size from 50 to 4000 for the CUHK-SYSU dataset to verify the influence of gallery size. And we report the results in terms of CMC top-1 and mAP on the CUHK-SYSU  dataset. As illustrated in Fig. \ref{GS}, as the size of the map library increases, the mAP, and CMC top-1 decrease. Since the PRW data set uses 6112 pictures in the complete set for testing, we randomly selected [2000, 3000, 4000, 5000] pictures for experiments in order to better verify the effectiveness of the model in real scenarios, and took the average value after many experiments. The resulting mAP and CMC top-1 are shown in Figure. \ref{PR}. Our method is superior to other methods in the PRW dataset, which proves the robustness of our model. And the decline of the mAP and CMC top-1 is smaller than that of other methods, which also proves that our previous analysis of DA-GSS is more suitable for real scenarios. %%没改

\begin{figure}[tbp!]
	\centering
	\includegraphics[width=\textwidth]{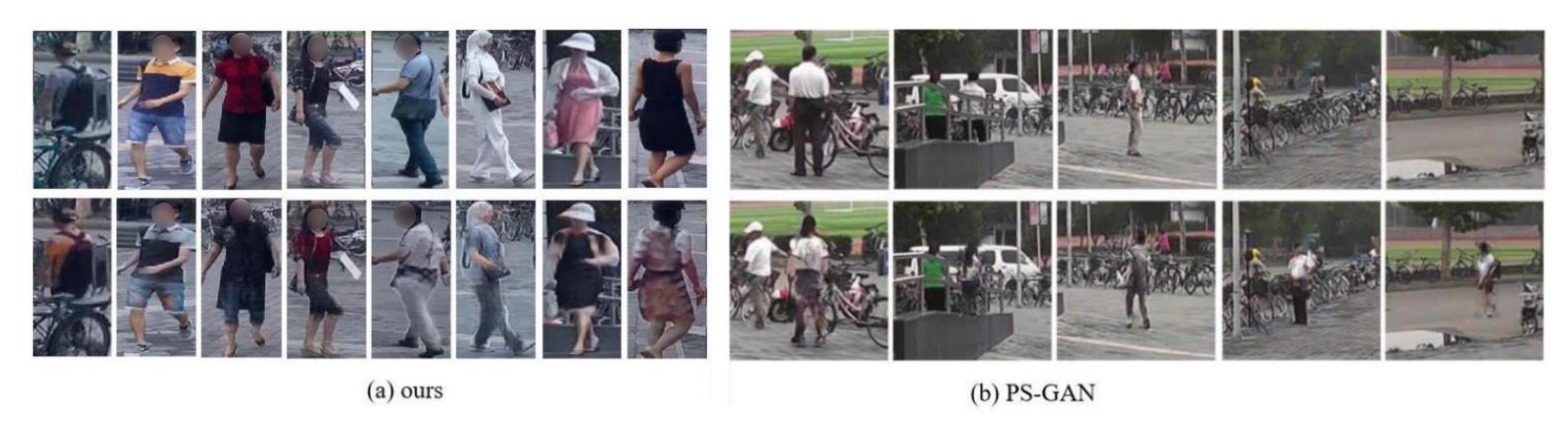} 
	\caption{Comparison of the generated and real images on PRW across PS-GAN \cite{yao2020gan}, and our approach. This figure is best viewed when zoom in.}
	\label{watch}
\end{figure}

Figure. \ref{it} shows the performance of our DA-GSS on the two datasets for each set of iterations([20k,40k,60k,80k,100k]). We can see that as the number of iterations increases, GAN can generate more vivid and realistic images. Among previous research works, PS-GAN\cite{yao2020gan} is the only method that introduces generative adversarial networks in the pedestrian search task. Unlike us, it uses GAN outside the pedestrian search framework to transfer people from one scene to another. We show the visualization results of both methods in Figure. \ref{watch}, where our DA-GSS generates realistic and diverse images.

\begin{table}[htbp!]
	\centering
	\caption{Comparison with two-step models on PRW, w.r.t. effectiveness and efficiency. }
	\label{Table6}
		\begin{tabular}{cccccc}
			\hline
			Methods     & GFLOPS & PRARMS & Time & map  & top-1 \\ \hline
			FRCNN+MAR   & 148    & 113.1M & 33   & 42.8 & 84.5  \\
			FRCNN+PNGAN & 110    & 78.5M  & 27   & 42.5 & 82.3  \\
			FRCNN+DGnet & 118    & 88.6M  & 29   & 50.2 & 89.4  \\
			Ours        & 126    & 90.2M  & 31   & \textbf{58.3} & \textbf{91.2}  \\ \hline
	\end{tabular}
	
\end{table}
To further assess the effectiveness and efficiency of this framework, we conducted a comparative analysis with other various two-step models, which first utilize a detector to localize pedestrians(Faster-RCNN), and then apply domain adaptation or generative adversarial network Re-identification methods \cite{yu2019unsupervised, qian2018pose, zheng2019joint} for person search. Table \ref{Table6} demonstrates that our two-step method surpasses other two-step models in performance while also exhibiting notable advantages in efficiency. Specifically, we employed the same backbone (ResNet-50) for all two-step models to ensure consistency in model complexity and runtime analysis. Consequently, they exhibit comparable FLOPS and running time during inference. Due to our model's ability to generate highly precise positive samples and learn more distinctive features, it achieves exceptional performance at a low computational cost. To assess the efficacy of the proposed DA-GSS model in this regard, we present our findings in Table \ref{Table4}. The substantial number of detection boxes generated by Faster-RCNN, including numerous background and other unmarked instances, led to inferior results. Nevertheless, despite generating fewer detection boxes, the AIDQ module's performance was not particularly noteworthy. Our analysis suggests that the significant presence of hard negative samples has a detrimental impact on the model's learning ability. The integration of the GAN synthesis module effectively resolves the issue of limited positive samples, and facilitates the model's ability to learn more distinctive features. Hence, the performance of person search is contingent not on the quantity but rather the quality of proposals.

\begin{figure}[tbp!]
	\centering
	\includegraphics[width=\textwidth]{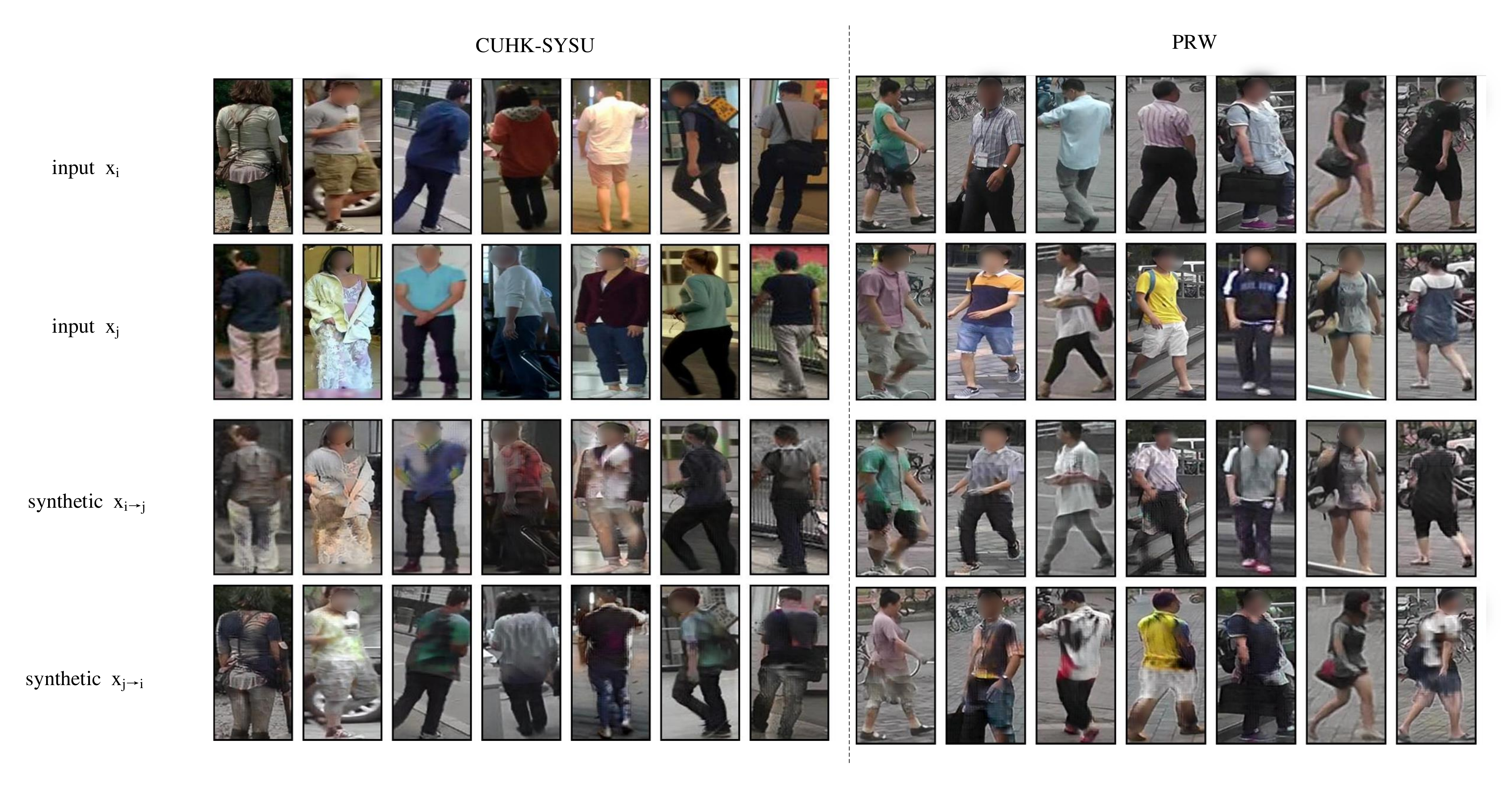} 
	\caption{Examples of images generated by swapping clothing structures within the two datasets.}
	\label{sys}
\end{figure}

Additionally, as indicated in Table \ref{Table5}, the performance of our model is significantly influenced by the choice of backbone. Utilizing ResNet-18 yields a MAP of 42.5\%, whereas incorporating ResNet-101 results in a substantial improvement to 58.7\%. However, the use of a deeper backbone network also results in significantly higher FLOPS (78G → 159G) and larger parameters (68.5M → 117.3M). Hence, selecting a more appropriate backbone network also demands thoughtful consideration. 

\begin{figure}[tbp!]
	\centering
	\includegraphics[width=\textwidth]{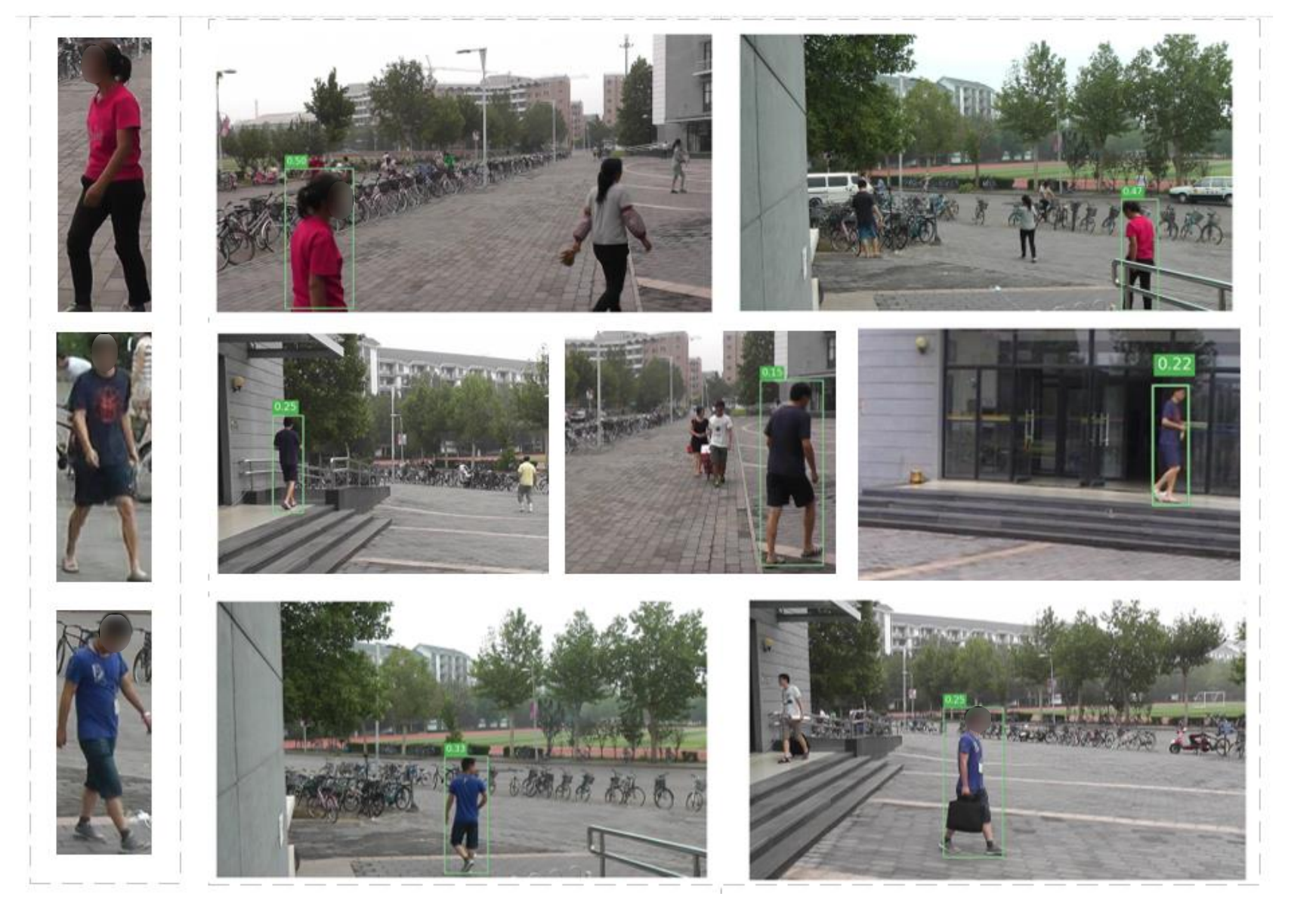} 
	\caption{Visualization results of person search. The left side represents pedestrians to be searched and the green bounding boxes represent search results.}
	\label{visual}
\end{figure}

\subsubsection{Visualization}
We showcase our generative results on two benchmarks in Figure. \ref{sys}, highlighting the consistent generation of realistic and diverse images by DA-GSS across various datasets. Furthermore, we visualize the search results on a challenging dataset to validate the effectiveness of DA-GSS. Figure. \ref{visual} illustrates the query image on the left and the corresponding gallery image displaying pedestrians with accurate matches on the right.

\begin{table}[tbp!]
	\centering
	\caption{Evaluating effectiveness of AIDQ model and our proposed DA-GSS on PRW. The number of ground truthboxes is 14907. }
	\label{Table4}
		\begin{tabular}{ccccc}
			\hline
			Methods      & Boxes Num & Precision & map  & top-1 \\ \hline
			Faster R-CNN & 30,597    & 95.2   & 39.4 & 84.3  \\
			AIDQ(ours)   & 18,873    & 96.8   & 37.2 & 77.1  \\
			DA-GSS(ours) & 31,132    & 96.8   &\textbf{58.3} &\textbf{91.2} \\ \hline
	\end{tabular}
\end{table}

 \begin{table}[]
	\centering
	\caption{Results on PRW with various backbones.  }
	\label{Table5}
		\begin{tabular}{ccccc}
			Methods    & GFLOPS & PRARMS  & map  & top-1 \\ \hline
			ResNet-18  & \textbf{78}     & \textbf{68.5M}   & 42.5 & 82.3  \\
			ResNet-34  & 103    & 81.0M   & 45.0 & 86.4  \\
			ResNet-50  & 126    & 90.2M   & 58.3 & 91.2  \\
			ResNet-101 & 159    & 117.3M  & \textbf{58.7} & \textbf{92.0}  \\ \hline
		\end{tabular}
\end{table}
% Please add the following required packages to your document preamble:
% \usepackage{multirow}

   \subsection{Ablation Study}
   To validate the effectiveness of each component of DA-GSS, we perform several ablation experiments on CUHK-SYSU and PRW datasets. And the gallery size is set to 100 for all experiments.

   We first discuss whether the improvement brought by DA-GSS comes from the positive samples provided by AIDQ or from more discriminative features provided by synthetic images. The result is shown in Table \ref{Table2}. As expected, the results obtained by using the AIDQ module and synthetic images method are much higher than those obtained by using alone or not. In addition, we also discover another key point that the PRW dataset is improved much more by the synthetic image than by the CUHK dataset, which proves that our model is more suitable for real camera scenes. For a dataset like CUHK, which contains multiple scenarios, our model still has much room for improvement. It is also the first time that we have introduced a generative countermeasure network into the person search framework to synthesize images. Although the effect is not particularly ideal, we hope it can enlighten future research work.
   \begin{table}[]
   	\centering
   	\caption{Performance of different components on two datasets. The ablation study about AIDQ module and Scene Synthesis module is in the upper block.} 
   	\label{Table2}
   	\begin{tabular}{cccccc}
   		\hline
   		\multirow{2}{*}{AIDQ} & \multirow{2}{*}{\begin{tabular}[c]{@{}c@{}}Scene\\Synthesis \end{tabular}} & \multicolumn{2}{c}{CUHK-SYSU}               & \multicolumn{2}{c}{PRW}                     \\ \cline{3-6} 
   		&                                                                           & mAP                 & top-1                & mAP                  & top-1                \\ \hline
   		&                                                                           & 78.2                 & 83.1                 & 35.8                 & 75.4                 \\
   		$\checkmark$         &                                                                          &84.6                      &88.3                      &37.2                      &77.1                      \\
   		&$\checkmark$                                                                           &83.9                      &87.5                      &50.8                      &88.3                      \\
   		$\checkmark$  &$\checkmark$                                                     &\textbf{87.6} &\textbf{93.5} &\textbf{58.3} &\textbf{91.2} \\ \hline
   	\end{tabular}
   \end{table}
   In addition, the learning of hard negative samples can better widen the distance between positive samples and negative samples in our AIDQ design. We adjust the lambda parameter to change the number of hard negative samples selected in Eq. \ref{4}. As can be seen from Figure. \ref{lambda}, our model is sensitive to this hyperparameter. When $\lambda < 0.4$ , the performance drops significantly due to ignoring too many simple samples. However, by choosing an appropriate hard negative ratio (i.e., 0.4 $\sim$ 0.8), our proposed strategy achieves significant improvement over the baseline method without hard negative sampling ($\lambda$ = 1). These results validate the effectiveness of our method.
   
   We also experimented with thresholds in detection, and different thresholds will output different numbers of images. We calculate the intersection over union(IoU) of the real frame and the detection frame, retain and crop the detection frame whose intersection ratio is greater than the threshold and pass it into the re-id task. We show the effect of different thresholds on experimental accuracy in Table \ref{Table3}. When the threshold is set to 0.6/0.5, the model achieves the best performance on CUHK-SYSU/PRW dataset.

\begin{table}[]
	\centering
	\caption{Top-1 and map with different threshold on CUHK-SYSU and PRW datasets. } 
	\label{Table3}
	\begin{tabular}{cccc}
		\hline
		Datasets                   & Threshold & map & top-1 \\ \hline
		\multirow{4}{*}{CUHK-SYSU} & 0.4       &86.5     &91.7       \\
		& 0.5       &87.1     &92.4       \\
		& 0.6       &\textbf{87.6}     &\textbf{93.5}       \\
		& 0.7       &87.5     &93.3       \\ \hline
		\multirow{4}{*}{PRW}       & 0.4       &58.1     &90.8       \\
		& 0.5       &\textbf{58.3}     &\textbf{91.2}       \\
		& 0.6       &58.2     &91.2       \\
		& 0.7       &57.9     &90.7       \\ \hline
	\end{tabular}
\end{table}
% Please add the following required packages to your document preamble:
% \usepackage{multirow}

\section{Conclusion}
In this paper, we noticed that the existing research work does not solve the cross-domain problem in person search very well. In real scenes, due to the influence of weather, different types of cameras, and other factors, the performance of some models dropped off a cliff. To address this issue, we propose a GAN-based Scene Synthesis framework for domain adaptive person search. Specifically, this is the first time that GAN has been introduced inside a pedestrian search framework. We design an Assisted-Identity Query Module (AIDQ) in the detection framework to provide positive images for the re-ID task. In addition, we also devise a GAN to generate high-quality cross-identity person images and enable the re-id model to learn more fine-grained and discriminative features through an online learning strategy. Extensive experiments on two widely used person search benchmarks, CUHK-SYSU and PRW, have shown that our method has achieved great performance on the PRW dataset. However, the cross-domain images generated in the CUHK-SYSU dataset containing street images and movie screenshots are not satisfactory. We hope that our research has some inspiration for future researchers
\section*{Acknowledge}
This work was supported in part by the National Natural Science Foundation of China Grant 62002041, Grant 62176037, the Liaoning Fundamental Research Funds for Universities Grant LJKQZ2021010, the Liaoning Doctoral Research Startup Fund Project Grant 2021-BS-075 and the Dalian Science and Technology Innovation Fund 2021JJ12GX028 and 2022JJ12GX019.

\bibliographystyle{elsarticle-num}
\bibliography{mybibfile}

\end{document}